\documentclass[twoside,11pt]{article}

\usepackage{jmlr2e}
\usepackage{listings}
\usepackage{bm}

\usepackage{pgfplots}
\usepackage{tikz}
\usepackage{makecell}

\usepackage{todonotes}

\usepackage{xcolor}
\definecolor{codegreen}{rgb}{0,0.6,0}
\definecolor{codegray}{rgb}{0.5,0.5,0.5}
\definecolor{codepurple}{rgb}{0.58,0,0.82}
\definecolor{backcolour}{rgb}{0.95,0.95,0.92}

\lstdefinestyle{mystyle}{
    backgroundcolor=\color{backcolour},   
    commentstyle=\color{codegreen},
    keywordstyle=\color{magenta},
    numberstyle=\tiny\color{codegray},
    stringstyle=\color{codepurple},
    basicstyle=\ttfamily\footnotesize,
    breakatwhitespace=false,         
    breaklines=true,                 
    captionpos=b,                    
    keepspaces=true,                 
    numbers=left,                    
    numbersep=5pt,                  
    showspaces=false,                
    showstringspaces=false,
    showtabs=false,                  
    tabsize=2
}

\ShortHeadings{MKLpy}{Lauriola and Aiolli}
\firstpageno{1}

\begin{document}

\title{MKLpy: a python-based framework for\\Multiple Kernel Learning}

\author{\name Ivano Lauriola \email ivano.lauriola@phd.unipd.it \\
        \name Fabio Aiolli \email aiolli@math.unipd.it \\
       \addr Department of Mathematics,
       University of Padova, 
       35121 Italy}

\editor{\dots}

\maketitle

\begin{abstract}
Multiple Kernel Learning is a recent and powerful paradigm to learn the kernel function from data.
In this paper, we introduce MKLpy, a python-based framework for Multiple Kernel Learning.
The library provides Multiple Kernel Learning algorithms for classification tasks, mechanisms to compute kernel functions for different data types, and evaluation strategies.
The library is meant to maximize the usability and to simplify the development of novel solutions.
\end{abstract}


\section{Introduction}
Kernel machines~\citep{shawe2004kernel} are a popular family of machine learning algorithms widely used in the literature to solve classification, regression, and clustering problems.
These methods comprise two parts, (i) a function, named kernel, defining the similarities between pairwise examples and (ii) a learning algorithm leveraging these similarities.
Recently, several methods have been proposed to directly learn the optimal kernel function from data, overcoming the limits of an expensive validation process.
One of the most popular and effective kernel learning frameworks is the
Multiple Kernel Learning~\citep{gonen2011multiple} (MKL), whose purpose is to learn the kernel as a principled combination of several base (or weak) kernels.
Nowadays, MKL is gaining popularity in different domains and, to the best of our knowledge, it is mainly applied to two categories of tasks, that are (i) information fusion, where users define a base kernel for each view or source (e.g. audio, video, and text) \citep{xu2013survey,6977360}, and (ii) deep kernel learning, where the goal is to create kernels with the deep structure that characterizes neural networks~\citep{donini2017learning,lauriola2020enhancing}.

Having said that, this paper introduces MKLpy: a python-based framework for MKL.
Specifically, we hereby describe the most important features and characteristics of the library, focusing on the main use-cases and the computational challenges addressed. 

\section{Features and use-cases}

\begin{figure}
    \centering
    \includegraphics[width=.85\linewidth]{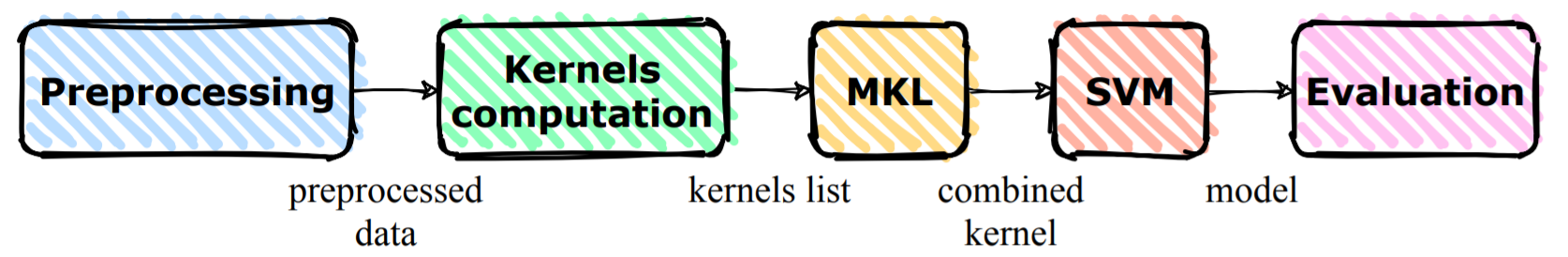}
    \caption{MKL pipeline. The intermediate output between states is also exposed.}
    \label{fig:pipeline}
\end{figure}

MKLpy encapsulates \emph{everything you need} to run MKL algorithms, to compute kernels, and to evaluate solutions, 
and it is specifically designed in accordance with two main principles, i.e. the usability and the extendability, covering different needs from user and developer perspectives.
On one hand, the library provides high-level APIs to easily apply MKL algorithms to the user's specific task. 
Inspired by the scikit-learn~\citep{pedregosa2011scikit} project, the public interface of MKL algorithms consists of simple primitives to train an algorithm \texttt{mkl.fit(\dots)} and to perform predictions \texttt{mkl.predict(\dots)}. 
On the other hand, MKLpy is designed to dissect the entire MKL pipeline, depicted in \figurename~\ref{fig:pipeline}, from the computation of base kernels to the inference phase, allowing a modular, incremental, and targeted development.
The main features of MKLpy can be divided into 3 categories:
\begin{description}
    \item[Kernel functions]: 
    MKLpy natively supports several kernel functions for (i) vectorial data, such as the popular homogeneous polynomial kernel (HPK), (ii) binary valued data, such as the conjunctive and disjunctive boolean kernels, and (iii) structured data, such as the $p$-spectrum and string kernels. 
    Additionally, MKLpy supports all the kernels from the scikit-learn package, which have the same public interface, and custom kernels.
    In the following snippet, we show a basic example of kernels computation.
    Specifically, we create a single monotone conjunctive kernel (\texttt{K}), and a list of kernels (\texttt{KL}) for MKL algorithms.
\begin{lstlisting}[language=Python]
 from MKLpy.metrics.pairwise import monotone_conjunctive_kernel as mck
 K  = mck(X)   #X is a scikit-like training matrix or Torch.Tensor
 KL = [mck(X, c=arity) for arity in range(1,6)]
\end{lstlisting}

    \item[MKL algorithms]: The core of the library is the implementation of various MKL algorithms, such as AverageMKL\footnote{AverageMKL is a simple wrapper computing the average of base kernels.}, EasyMKL~\citep{aiolli2015easymkl},  
    R-MKL~\citep{do2009margin}, 
    and GRAM~\citep{lauriola2017radius}. The algorithms are available with specialized wrappers exposing the same public interface of scikit-learn predictors. The wrappers automatically encapsulate (i) the kernels combination mechanism, (ii) the training leveraging the resulting kernel, and (ii) the prediction mechanism.
\begin{lstlisting}[language=Python]
 from MKLpy.algorithms import AverageMKL, EasyMKL
 mkl = AverageMKL()  #create a MKL instance
 mkl = mkl.fit(KLtr, Ytr)    #train the model
 predictions = mkl.predict(KLte)
\end{lstlisting}
    We also provide several customizations in the pipeline to have more control over execution.
    For instance, users may perform only the combination step, inspect the weights, and specify the kernel method dealing with the combination, such as the popular SVM,  KOMD~\citep{aiolli2008kernel}, or any scikit-compliant kernel machine.
    
    \item[Evaluation]: An important step in kernel learning is the evaluation of the learned kernels. MKLpy offers several criteria and tools to evaluate the quality of a kernel, such as the radius of the Minimum Enclosing Ball~\citep{gai2010learning,chung2003radius} (MEB) containig data in the kernel space, the margin, the alignment between kernels~\citep{kandola2002optimizing}, or the empirical complexity (Spectral Ratio)~\citep{donini2017learning}. 
\begin{lstlisting}[language=Python]
 from MKLpy import metrics
 margin = metrics.margin(K,Y)    #margin between classes
 radius = metrics.radius(K)      #radius of MEB
 SR     = metrics.spectral_ratio(K, norm=True)   #empirical complexity
\end{lstlisting}
    
    Additionally, the user may leverage tools from the scikit-learn package to evaluate the performances in terms of accuracy, AUC, and other metrics.
\end{description}




\section{Computational efficiency}

Besides scikit-learn, MKLpy relies on multiple scientific libraries, namely numpy~\citep{walt2011numpy}, CVXOPT~\citep{andersen2013cvxopt}, and PyTorch~\citep{NEURIPS2019_9015}. These high-level libraries leverage, in turn, low-level routines, provided by BLAS~\citep{blackford2002updated} and LAPACK~\citep{laug}, that execute the most of operations. 
To speed up the computation, we strongly suggest to correctly install, configure, and link the most appropriate configuration and external libraries before installing MKLpy, according to the system architecture.
Furthermore, we recommend running the code on a multi-core machine with the AVX2 instructions set.

The main bottleneck of MKL algorithms preventing to scale with large datasets is memory consumption. 
In order to improve the efficiency, most of the optimization engines used in this library require to keep the whole kernels list in memory during training.
In order to make the resource allocation tractable and comparable to a single SVM, we provide specialized kernels generators that perform a lazy computation of kernels matrices, making them available on-the-fly only when they are needed during optimization.
The generators are designed to reduce as much as possible the required memory and to speed up the kernels computation, when possible, with specific caching mechanisms. 
\begin{lstlisting}[language=Python]
 from MKLpy.generators import HPK_generator #homogeneous polynomial kernels
 KLtr = HPK_generator(Xtr, degrees=range(1,21))
 mkl  = EasyMKL.fit(KLtr, Ytr)
\end{lstlisting}

In the Table~\ref{tab:generators} we show the time and memory consumption required by EasyMKL with 20 HPKs applied to two datasets, Madelon and Phishing\footnote{The datasets are freely available on \url{https://www.csie.ntu.edu.tw/~cjlin/libsvmtools/datasets/}}.
Specifically, we show the resources usage when dealing with explicit lists of kernels and MKLpy generators. 
The values include the kernels computation and model training.
What is striking from the table is the huge reduction of memory consumption, with the average of 3.7 times. 


\begin{table}[htb]
    \small
    \centering
    \begin{tabular}{l|c|c|c|c|c}
        \hline
        Dataset & examples & features & list & generator w. cache & generator \\
        \hline
        Madelon & 6000 & 5000 & \makecell{24.4 s / 7.3 GB} & \makecell{28.0 s / 2.6 GB} & \makecell{60.5 s / 2.3 GB} \\
        \hline
        Phishing & 11055 & 68 & \makecell{115.7 s / 23.9 GB} & \makecell{120.4 s / 6.2 GB} & \makecell{126.8 s / 5.7 GB} \\
        \hline
    \end{tabular}
    \caption{Computational time and memory usage of EasyMKL when trained with explicit kernels list and MKLpy generators.}
    \label{tab:generators}
    \vspace{-1.5em}
\end{table}

\section{Guidelines and best practices}

The development of novel MKL algorithms may often be tricky and brimful of smokescreens, and there are several details and practices that need to be taken into account during the development and evaluation processes. 
%
To this end, we hereby describe a useful set of guidelines and best practices to develop and to properly evaluate novel MKL algorithms leveraging this library. 
These guidelines are mainly inspired by our research experience and by the numerous problems we encountered during the development of MKLpy, and they are listed in the following:
\begin{itemize}
    \item MKL algorithms leverage the concept for which the combination of multiple representations improve the single representation. This aspect is fundamental but not obvious. Indeed, different representations may introduce noise or useless information. To this end, a comparison of the combined kernel against the single base kernels may certify the quality of the MKL solution.
    \item Despite its simplicity, several existing MKL algorithms do not even perform better than the simple average of base kernels. 
    However, this baseline is rarely considered in the assessment process.  Thus, we strongly encourage to use this baseline when evaluating MKL solutions. 
    \item The combination weights provide useful insights concerning the sparsity of the solution and the contribution of base kernels, and their analysis prevents the development a trivial solution. For instance, the algorithm may assign 0 weights to all kernels but one. In this case, the MKL can be used for feature selection or hyper-parameter tuning~\citep{massimo2016hyper}.
    \item The weight assigned to each kernel depends on the norm of the latter. If we scale up the norm of a base kernel, its contribution in the combination may scale up as well\footnote{This happens, for instance, with EasyMKL.} and the kernel may receive a high weight just because it has a high norm.
    This aspect may provide a sub-optimal solution in some cases, and the resulting weights vector may be misleading.
    To reduce this problem, we suggest normalizing base kernels. 
\end{itemize}
To this end, MKLpy provides multiple callbacks and tools to visualize the weights vector over iterations, to monitor the training objective, and to evaluate validation metrics.

\section{Conclusions}

This paper introduces MKLpy, a python-based framework for Multiple Kernel Learning (MKL). 
The framework encapsulates everything you need to run and to evaluate MKL algorithms. 
MKLpy is specifically designed to maximize the usability, and to support the development of novel algorithms and functionalities.
Here, we only exposed the main concepts of the library, without providing a deep dive into specific functionalities.
Exhaustive examples covering a plethora of use-cases and implementation details for developers are available on our documentation\footnote{Documentation: \url{https://mklpy.readthedocs.io/en/latest/}}.
The code of MKLpy is currently hosted on GitHub\footnote{GitHub: \url{https://github.com/IvanoLauriola/MKLpy}}.



\bibliography{bib}

\end{document}